# Similar Document Template Matching Algorithm

Bommareddy Revanth Srinivasa Reddy
School of Computer Science and Engineering
Vellore Institute of Technology
Chennai, India

Harshitha Yenigalla
School of Computer Science and Engineering
Vellore Institute of Technology
Chennai, India

Batta Venkata Rahul
School of Computer Science and Engineering
Vellore Institute of Technology
Chennai, India

N Hemanth Raju
School of Computer Science and Engineering
Vellore Institute of Technology
Chennai, India

*Abstract*—:**This research presents a thorough approach to medical document verification that incorporates cutting-edge methods for fraud detection, template extraction, and comparison. The process starts with the extraction of the template using advanced region-of-interest (ROI) techniques that include edge identification and contour analysis. By using adaptive thresholding and morphological operations, pre-processing procedures guarantee template clarity. By using advanced feature matching with key points and descriptors, the template comparison algorithm improves robustness. The SSIM computation and OCR for textual information extraction are used in fraud detection. By quantifying structural similarity, the SSIM facilitates the identification of possible matches. Critical areas such as patient details, provider information, and billing amounts are the focus of OCR. Reliable fraud detection is ensured by confidence thresholding and comparing the extracted data with a reference dataset. Flexible parameters allow the system to adapt dynamically to different document layouts. This methodology addresses complexity in template extraction, comparison, fraud detection, and adaptability to different document structures, offering a strong approach to medical document verification.**

## I. INTRODUCTION:

A crucial component of several fields, such as document analysis, data mining, and information retrieval, is document template recognition. Finding comparable document templates is essential for expediting procedures and stopping false claims, which can lead to large financial losses and undermine system trust. This introduction explores the development of template recognition methods, focusing on recent developments in deep learning, shape matching, and graph-based representations as well as conventional methods.

Pixel-wise or feature-based comparisons have always been the foundation of traditional template matching. It was usual practice to use methods like structural similarity indices and normalized cross-correlation. These techniques demonstrated susceptibility to noise, rotation, and size fluctuation, despite being effective in some situations. Their inability to adjust to different document layouts was a result of their reliance on pre-established norms and patterns. In order to overcome the shortcomings of conventional methodologies, feature-based approaches became more popular. But shifts in perspective and attitude brought difficulties. Template matching algorithms, which are divided into supervised and unsupervised learning approaches, underwent a dramatic change with the introduction of machine learning.

Under supervised learning, labelled data was used to learn template patterns through the use of algorithms such as decision trees and support vector machines. This improved flexibility to a variety of document structures. Unsupervised methods like topic modelling and clustering have become popular for classifying related documents without labelled data. Deep learning techniques, particularly Convolutional Neural Networks (CNNs) and Recurrent Neural Networks (RNNs), have become more popular in recent years.

Managing layout differences in documents, scaling for big datasets, and adapting to real-world circumstances where documents may have dynamic structures are additional challenges. Other factors to take into account are the interpretability of deep learning models and the requirement for labelled training data. Applications for similar document template recognition can be found in a variety of fields, including law, finance, and healthcare. Effective template identification simplifies document management in legal environments, supporting legal research and tasks like contract analysis. The effectiveness of information extraction and document management procedures is increased thanks to these algorithms. Additionally, they are essential in identifying fake document templates, protecting a variety of sectors from large financial losses. The suggested methodology uses a comprehensive strategy that includes template extraction, template comparison, structural similarity and optical character recognition (OCR) fraud detection to meet the issues of fraudulent document identification. The process starts with sophisticated ROI (region-of-interest) techniques for extracting templates.

The software uses image processing methods like edge identification and contour analysis to identify key areas in medical documents that have patient information, provider information, and billing amounts. These areas have been identified and are intentionally divided as possible templates. Before extraction, documents go through a number of pre-processing steps to improve the accuracy of template identification. Gaussian blurring lowers noise, adaptive thresholding increases contrast, and morphological operations are used to smooth and fine-tune image structures. Together, these pre-processing techniques yield crisp, well-defined images that provide a solid basis for the other processing stages. Key point and descriptor-based advanced feature matching techniques are employed by the template comparison algorithm. Recognized in the template and sample photos are important details like corners and distinct sections.

The first step in the multi-step process of detecting fraud is calculating the Structural Similarity Index (SSIM) between the sample images and the template's grayscale representations. A numerical measure of structural information similarity is offered by the SSIM. In later stages of fraud detection, the SSIM value serves as a standard that helps establish the legitimacy of the reviewed medical record. Optical character recognition (OCR) is integrated into the methodology to extract text from medical records. OCR technology converts machine-readable text from images of written or printed text. Text localization is a step in the OCR process where template matching information identifies regions of interest that are most likely to have textual content. Text extraction extracts textual material, such as patient names, addresses, and other relevant information, by using OCR techniques.

After being extracted, the textual data is compared to a reference dataset, which provides reliable and accurate data as a basis. The dataset includes accurate patient data, provider data, and hospital-sourced billing records. A methodical comparison is carried out to look for discrepancies or inconsistencies between the information that was retrieved and the reference dataset. Any discrepancies are reported as possible indicators of false claims. A confidence thresholding strategy is used to increase the accuracy of fraud detection. A confidence score is assigned to each stage of the attribute comparison and OCR process. After template matching and OCR results are combined, the confidence ratings are compared to a preset threshold. A document is only considered possibly fraudulent if the cumulative confidence is higher than this level. This thresholding system strikes a balance in the approach by ensuring sensitivity to potential fraud while minimizing false positives. The methodology integrates flexibility by means of modifiable parameters. Depending on the characteristics of the input documents, parameters like the matching criteria and feature extraction parameters are dynamically altered. This adaptability takes into account the inherent variability in document structures and guarantees consistent performance across a broad range of medical document layouts.

The recognition of document templates has become an essential component in many fields, tackling problems related to data mining, document analysis, and information retrieval. The possibility of identical document templates with minute modifications, which present opportunities for fraudulent claims that can result in significant financial losses and erode trust in systems, highlights the importance of this task. In addition to helping to increase the effectiveness of information extraction and document management procedures, these algorithms are essential in identifying fraudulent document templates and averting significant financial losses across a range of industries. The detailed methodology and implementation that follows provides an organized way to deal with the difficulties associated with fraudulent document detection. The introductory section establishes the foundation for a thorough comprehension of the dynamic field of template recognition and its pragmatic uses.

## II. LITERATURE SURVEY:

Document template recognition is an important task with applications in various domains, such as information retrieval, data mining, and document analysis. There is a chance of identical document templates with minute modifications. These kinds of fraudulent claims can lead to significant financial losses and erode the trust in their systems. Recognizing similar document templates involves identifying patterns and structures within documents to categorise them based on their underlying templates. This literature survey explores the advancements in similar document template recognition, focusing on key methodologies, challenges, and applications.

2.1. Traditional Template Matching Techniques:

Traditional template matching often involved methods based on pixel-wise or feature-based comparisons. Techniques such as normalised cross-correlation and structural similarity indices were commonly used. These methods are effective in certain scenarios but they are sensitive to size variation, rotation and noise. These approaches focused on predefined rules and patterns, making them limited in adaptability to diverse document layouts.

## 2.2 Feature-based Approaches:

Feature-based methods have gained popularity for their ability to capture distinctive elements within documents. Keypoint detectors, such as SIFT (Scale-Invariant Feature Transform) and SURF (Speeded-Up Robust Features), have been used to extract discriminative features for template matching. These techniques enable the extraction of meaningful features that contribute to a document's template categorization. However, these methods may struggle with changes in orientation and perspective.

## 2.3 Machine Learning-based Matching:

With the increased usage of machine learning, template matching algorithms have been divided into supervised and unsupervised learning techniques. The rise of machine learning techniques has marked a significant advancement in document template recognition. Supervised learning algorithms/techniques, including Support Vector Machines (SVM) and decision trees, have been applied to learn template patterns from labelled data, enhancing the adaptability of matching algorithms to diverse document structures. Unsupervised techniques such as clustering and topic modelling have also gained popularity for grouping similar documents without labelled information.

## 2.4 Deep Learning Approaches:

Recent years have seen the rise of deep learning approaches, particularly Convolutional Neural Networks (CNNs) and Recurrent Neural Networks (RNNs). CNNs excel in extracting special features from images, making them suitable for tasks involving document layout analysis. RNNs, on the other hand, are adept at modelling sequential dependencies, which is crucial for recognizing templates in text-heavy documents, leading to improved template recognition accuracy.

## 2.5 Shape Matching and Graph-based Representations:

Recent advancements in shape matching and graph-based representations have contributed to more powerful and advanced template matching. These techniques model the document structures as graphs and leverage graph matching algorithms have shown its advancement in capturing complex relationships between document elements, improving matching accuracy in scenarios where traditional methods may fall short.

Despite the advancements, there are many challenges that persist in this field. Variability in document layouts, handling dynamic templates, and managing noise in data are ongoing concerns. Creating comprehensive labelled datasets for training remains a challenge, impacting the performance of supervised learning models. Challenges in similar document template matching include handling variations in document layouts, addressing scalability issues for large datasets, and adapting to real-world scenarios where documents may exhibit dynamic structures. The interpretability of deep learning models and the need for labelled training data are also noteworthy considerations.

Similar document template recognition finds applications across various industries, including legal, finance, and healthcare. In legal settings, for instance, efficient template recognition streamlines document management, aiding in tasks such as contract analysis and legal research. These algorithms contribute to improved efficiency in document management and information extraction processes. These can detect the fraud document templates and can protect major financial losses in various industries.

Yang et al. proposed a hybrid matching method for document image template recognition, combining local features and global features to improve matching accuracy [3]. The proposed algorithm demonstrates superior performance compared to traditional matching methods. However, they may be computationally expensive due to the combination of multiple feature extraction methods.

Liu et al. introduced a robust document template matching method based on SIFT (Scale-Invariant Feature Transform) features, which are invariant to scale and rotation [4]. The matching method utilises SIFT features, which are invariant to scale and rotation, for robust matching.

This achieves high matching accuracy and robustness against noise and distortions. However, Scale-Invariant Feature Transform (SIFT) feature extraction can be computationally expensive for large document images and cannot be used for large scaling.

Deng et al. proposed a template matching approach for document image classification, utilising a combination of correlation matching and structural matching [5]. The proposed method demonstrates promising performance in classifying various types of document images. It uses a combination of correlation matching and structural matching for improved classification, such that different types of document images can be classified. However, it may struggle with complex document layouts with significant variations in structure.

Wu et al. proposed a context-aware document template matching method that incorporates contextual information into the matching process to improve accuracy [6]. It has shown superior performance which can handle variations in document layouts and enhances the matching performance. However, extracting contextual information can be challenging for documents with complex layouts or noisy content.

Lu et al. introduces a hierarchical document template matching approach based on graph matching [7]. The proposed method utilises graph structures to represent document templates and their relationships, enabling efficient matching and recognition. It utilises graph structures to represent document templates and their relationships for efficient matching. However, Graph matching can be computationally expensive for large and complex graph structures.

Li et al. proposed a deep template matching method for document image classification, employing deep learning techniques to extract and match features from document images [8]. It achieves high classification accuracy and shows superior performance than Traditional Template matching methods. However, it requires training deep learning models, which can be time-consuming and computationally expensive.

## III. METHODOLOGY AND IMPLEMENTATION

### 3.1 Template Extraction

#### 3.1.1 Mechanism for Template Extraction

Advanced region-of-interest (ROI) approaches are used in the implementation of the template extraction procedure. The programme recognises important areas inside medical papers by combining a number of image processing techniques, such as contour analysis and edge identification. These sections, which contain vital data including patient specifics, provider details, and bill amounts, are purposefully separated as possible templates.

#### 3.1.2 Pre-processing Steps

To guarantee the best possible template identification accuracy, documents go through a rigorous sequence of pre-processing stages before they are extracted as templates. These procedures include morphological operations to smooth and fine-tune picture structures, Gaussian blurring for noise reduction, and adaptive thresholding to boost contrast. Together, these pre-processing procedures help produce clear, well-defined pictures that serve as a strong basis for further processing phases.

### 3.2 Template Comparison

#### 3.2.1 Algorithm for Template Comparison

Advanced feature matching approaches based on key points and descriptors are used by the template comparison algorithm. In both the template and sample photos, key points such as corners and distinguishing sections are recognised. Descriptors, which describe local characteristics around key points, are then compared using advanced algorithms like the Scale-Invariant Feature Transform (SIFT) or Speeded Up Robust characteristics (SURF). This method considerably improves the robustness of template matching, especially when coping with rotation, scaling, and lighting fluctuations.

#### 3.2.2 Techniques for Accounting Variations

The template comparison technique uses histogram-based analysis to adjust for variances in design components and content. The method discovers similarities in colour patterns and textures by comparing histograms of colour distribution within the template and sample pictures. This strategy is especially useful when there are small changes in document design components.

### 3.3 Fraud Detection

#### 3.3.1 Structural Similarity Index (SSIM) Computation

Fraud detection is a multi-step process that ensures reliable identification of potentially fraudulent claims. The Structural Similarity Index (SSIM) between the grayscale representations of the template and sample images is computed first. The Structural Similarity Index (SSIM) is a statistic that measures how similar two photographs are. It assesses picture structure information, brightness, and contrast, offering a comprehensive assessment of similarity. A result of 1 shows that the photos are perfectly matched. The SSIM value ranges from -1 to 1, with 1 denoting identical pictures. The SSIM formula is seen in Figure 1, emphasising its function in assessing the structural similarity between grayscale representations of the template and sample pictures. This index acts as a quantitative metric, allowing the system to locate prospective matches based on structural information similarity. A greater SSIM implies a stronger likeness, whereas a lower SSIM may indicate conflicts or changes in the document's content.

The SSIM value is used as a criterion in the succeeding phases of fraud detection to determine whether a probable match has been found. Furthermore, the SSIM helps to the total trust evaluation by assisting in determining the validity of the examined medical document.

$$SSIM(x, y) = \frac{(2\mu_x\mu_y + c_1)(2\sigma_{xy} + c_2)}{(\mu_x^2 + \mu_y^2 + c_1)(\sigma_x^2 + \sigma_y^2 + c_2)}$$

Where,

| | | |
|---|---|---|
| $\mu_x$ | = | the picture sample mean of x |
| $\mu_y$ | = | the picture sample mean of y |
| $\sigma_{xy}$ | = | the covariance of x and y |
| $\sigma^2_x$ | = | the variance of x |
| $\sigma^2_y$ | = | the variance of y |
| c1, c2 | = | two variables to stabilize the division with weak denominator |

Secondly, optical character recognition (OCR) is used to extract textual information, with a specific emphasis on consumer details. This step aims to scrutinize the contents of medical documents, focusing on patient information, provider details, and billing amounts.

#### 3.3.2 OCR for Textual Information Extraction

Optical character recognition (OCR) is a critical technology for extracting textual information from medical records. OCR technology turns written or printed text from pictures into machine-readable text. OCR is used in this approach to identify areas of interest (ROI) during the template extraction step.

The OCR process includes the following steps:

a. Text Localization: The algorithm employs template matching information to pinpoint locations of interest within the sample picture. These areas are likely to include textual content.

b. Text Extraction: In this approach, OCR techniques, such as those supplied by the 'easyocr' library, are applied to the indicated regions to extract textual material. This contains patient names, addresses, and other pertinent information.

c. Confidence Scores: During OCR, each recognised text segment is awarded a confidence score that reflects the algorithm's confidence in the recognition's correctness. This score is determined by criteria such as the text's clarity and the algorithm's internal confidence measures.

3.3.3 Comparison with Reference Dataset

After extracting the textual information, the programme compares the resulting characteristics to a reference dataset. This dataset provides as a foundation of authentic and correct data. A thorough collection of valid patient details, provider information, and billing records acquired from hospitals may be included in the reference dataset. The retrieved attributes, such as patient names, addresses, and billing amounts, are compared to the reference dataset in a methodical manner. Inconsistencies or deviations between the retrieved information and the reference dataset are reported as potential signs of fraudulent claims.

3.3.4 Confidence Thresholding

A confidence thresholding approach is used to improve the reliability of fraud detection. A confidence score is applied to each phase of the OCR process and attribute comparison. The combined confidence ratings from template matching and OCR results are compared against a threshold. A document is identified as possibly fake only when the cumulative confidence reaches this threshold. This thresholding system maintains sensitivity to possible fraud while minimising false positives, ensuring a balanced approach.

The system improves its capacity to identify potentially fraudulent claims by assessing both the structural and textual components of medical papers by incorporating OCR technology and reference dataset comparison.

3.4 Flexibility

3.4.1 Adaptive Parameters

The technique includes adjustable settings to increase flexibility. These parameters, which include matching criteria and feature extraction parameters, are dynamically changed based on the input documents' properties. This versatility provides dependable performance across a wide range of medical document layouts.

IV. RESULTS WITH DISCUSSION

The algorithm performs template matching to identify a document type based on predefined templates, checks for potential fraud by comparing the structural similarity using SSIM between the sample document and the best-matched template, and then uses OCR (Optical Character Recognition) to extract text from the sample document and check if certain attributes are present in the dataset given.

4.1 Fraud Detection

4.1.1 Potential Fraud Cases

The algorithm helps in finding the highest matching score of the sample with the template. If the highest matching score is above a threshold (0.6), it considers the document a potential match with a specific template.

After identifying the best-matched template, the code calculates the Structural Similarity Index (SSI) between the template and the sample image using the SSIM function from the skimage. metrics module. The SSI measures the structural similarity between two images. If the SSI is below a certain threshold (0.8), it indicates a significant difference between the template and the sample image. If the SSI is below the threshold, the code prints "Fraud," suggesting that the sample document significantly differs from the expected structure.

In Fig.1, we can observe that as the score is below the threshold of 0.8, it is identified as Potential fraud. This could be due to minor structural changes or alignment in the document provided, warning the institution that it could potentially be fraudulent.

```
Drive already mounted at /content/drive; to attempt to forcibly remount, call drive.mou
Best Template: doc.png, Matching Score: 0.6566147804260254
Potential Fraud
```

**Fig.1 Potential Fraud document case**

4.1.2 Potential Fraud Cases: Error in data

The code utilizes EasyOCR to extract text from the sample image after the template matching and SSI analysis. It converts the extracted text to lowercase and checks if certain attributes (text patterns) are present in a dataset. If all the required attributes are found in the detected text, it prints "REAL DOCUMENT," indicating that the document is valid. Otherwise, it prints "Error in data: Potential Fraud."

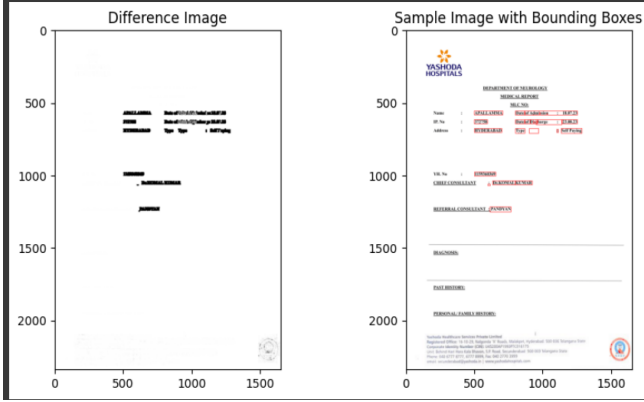

**Fig.2 Small section of the dataset**

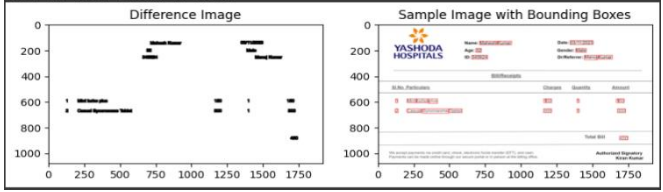

**Fig.3 Error in data: Potential fraud in medical report**

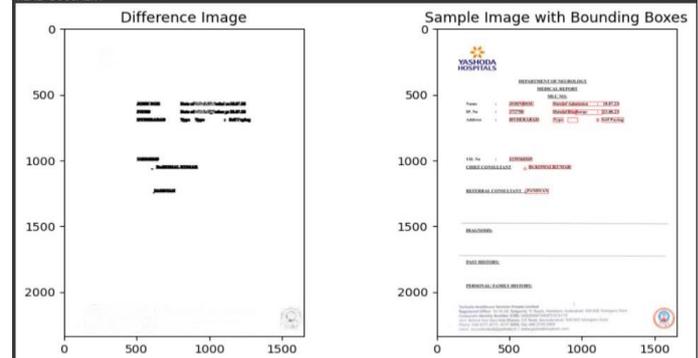

**Fig.4 Error in data: Potential fraud in bill**

In Fig.2 a snippet of the dataset used is given and Fig.3 a case of potential fraud case due to error in data comparison is shown. In Fig.4 that of a bill is shown.

4.1.3 Fraudulent Cases

If the SSI is below a certain threshold (0.8), it indicates a significant difference between the template and the sample image. The sample document significantly differs from the expected structure so the according to the algorithm this is considered Fraudulent.

In Fig.5 the logo is relocated and there are changes in the positioning and alignment of the boxes and text style and size. The matching score, and difference score indicate this case to be fraudulent, as indicated by the bounding boxes.

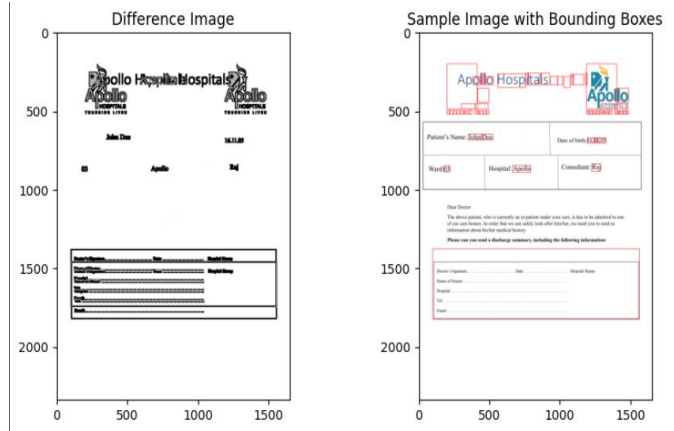

**Fig.5 Fraudulent Case**

4.1.4 Valid or Real Documents

In the last part of this paper, the detection of true and real documents is done based on the structural similarities done on the basis of threshold values and on checking for the data in the dataset, if proved true, "REAL DOCUMENT" is printed indicating that the document is real and isn't subjected to any form of fraud.

In Fig.7, an example of a real document along with its data base in Fig. 6.

| STONE | 575 | VIZAG | 06-07-2022 | 08-07-2022 |
| TONY | 576 | KODUMUR | 07-07-2022 | 09-07-2022 |
| MARK | 577 | NR.PETA | 08-07-2022 | 10-07-2022 |
| JOHN DOE | 372758 | HYDERABAD | 18.07.23 | 23.08.23 |

**Fig.6 Snippet of Database**

**Fig.7 True Document**

### 4.1.5 Architectural Diagram of the Proposed Approach

The primary goal of the script is to identify potential fraud by comparing a sample image with a set of template images and analyzing the structural similarity while performing OCR on the detected regions. The script starts by importing necessary libraries such as OpenCV, NumPy, EasyOCR, and others. Utilizes EasyOCR for text recognition. Overlays bounding boxes and recognized text on the image. Verifies if specific attributes are present in a dataset loaded from a CSV file. Performs template matching, structural similarity analysis, and OCR to detect potential fraud. If a high-quality match is found, the script calculates structural similarity to analyze image resemblance and uses OCR to extract text from the sample image. It checks the structural similarity score and, if below a threshold, flags the image as potential fraud. It then proceeds to check the detected text against a dataset to further validate potential fraud.

If potential fraud is detected, the script extracts text using OCR and checks if specific attributes are present in the detected text. It compares these attributes with a dataset loaded from a CSV file. The script provides output messages indicating whether a match was found, if potential fraud is detected, and whether the detected attributes match those in the dataset. It includes visual representations of the difference image and the sample image with bounding boxes for clearer understanding.

This architecture overview summarizes the key components and steps of the script, emphasizing its fraud detection capabilities through a combination of template matching, image analysis, and OCR as shown in Fig.9.

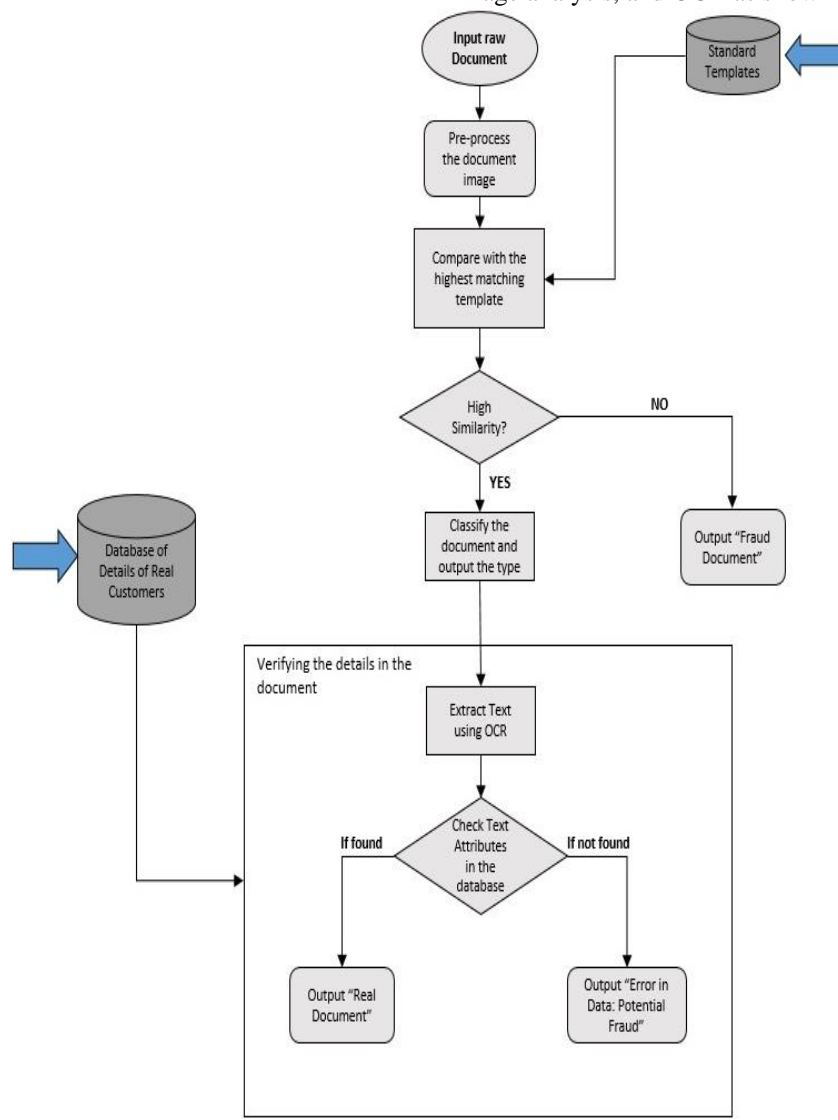

**Fig 8. Architectural diagram**

## V. CONCLUSION:

In conclusion, the literature survey underscores the dynamic evolution of similar document template recognition, encompassing a spectrum of techniques from traditional methods to state-of-the-art deep learning approaches. The surveyed literature highlights the significance of this field in various industries, showcasing its role in streamlining document management in legal settings, aiding in financial data analysis, and contributing to fraud detection mechanisms. While advancements in shape matching, graph-based representations, and deep learning have propelled the accuracy of template matching, challenges such as handling variations in document layouts persist. The need for comprehensive labelled datasets, scalability for large datasets, and adaptability to real-world scenarios emerge as ongoing concerns. The references provided in the survey offer a comprehensive overview of the field, drawing from seminal works that have shaped the trajectory of similar document template recognition research.